\documentclass{edm_article}
\usepackage{graphicx}%
\usepackage{multirow}%
\usepackage{amsmath,amssymb,amsfonts}%
\usepackage{mathrsfs}%
\usepackage[title]{appendix}%
\usepackage{xcolor}%
\usepackage{textcomp}%
\usepackage{manyfoot}%
\usepackage{booktabs}%
\usepackage{algorithm}%
\usepackage{algorithmicx}%
\usepackage{algpseudocode}%
\usepackage{listings}%
\usepackage{subcaption}
\usepackage{flushend}

\begin{document}

\title{Generating Feedback-Ladders for Logical Errors in Programming using Large Language Models}

%
%
%
%

\numberofauthors{2} 
%
\author{
%
%
\alignauthor
Hasnain Heickal\\
       \affaddr{University of Massachusetts Amherst}\\
       \affaddr{Amherst, MA, USA}\\
       \email{hheickal@cs.umass.edu}
\alignauthor
Andrew Lan\\
       \affaddr{University of Massachusetts Amherst}\\
       \affaddr{Amherst, MA, USA}\\
       \email{andrewlan@cs.umass.edu}
}

\maketitle              

\begin{abstract}
In feedback generation for logical errors in programming assignments, large language model (LLM)-based methods have shown great promise. These methods ask the LLM to generate feedback given the problem statement and a student’s (buggy) submission. There are several issues with these types of methods. First, the generated feedback messages are often too direct in revealing the error in the submission and thus diminish valuable opportunities for the student to learn. Second, they do not consider the student’s learning context, i.e., their previous submissions, current knowledge, etc. Third, they are not layered since existing methods use a single, shared prompt for all student submissions. In this paper, we explore using LLMs to generate a ``feedback-ladder'', i.e., multiple levels of feedback for the same problem-submission pair. We evaluate the quality of the generated feedback-ladder via a user study with students, educators, and researchers. We have observed diminishing effectiveness for higher-level feedback and higher-scoring submissions overall in the study. In practice, our method enables teachers to select an appropriate level of feedback to show to a student based on their personal learning context, or in a progressive manner to go more detailed if a higher-level feedback fails to correct the student’s error. \footnote{Published on the 17th Educational Data Mining (EDM) 2024}
\end{abstract}

\begin{table*}[h!]
\begin{tabular}{c||c}
\hline \hline
\textbf{Prompt} & \textbf{Response} \\
\hline 
\begin{minipage}[t]{.64\linewidth}
\vspace{5pt}
There can be different levels of feedback for a student who is trying to solve a programming assignment. Below we describe each level.

\textit{Level 0:}
Just the correct or incorrect verdict for the code.

\textit{Level 1:}
Giving a test case where the code fails. The test case contains just input, expected output and the code output. No explanations. 

\textit{Level 2:}
A high-level explanation of why the code failed in the test case. No mention of how to modify the code. 

\textit{Level 3:}
A high-level suggestion about the location in the code where you should make the changes. 

\textit{Level 4:}
Suggestion in actual programming language how to change the code to get the correct solution. Just the statements where change is necessary are mentioned. The full solution code is never given. 

For the given problem and code, generate feedback for each of these levels. When generating test cases make sure the generated test case falls inside the valid range.   
\\
\textbf{Problem:} Write a function in Java that implements the following logic: Given 2 ints, a and b, return their sum. However, sums in the range 10..19 inclusive, are forbidden, so in that case just return 20.\\
\textbf{Code:}
\begin{lstlisting}[breaklines, linewidth=\textwidth, basicstyle=\ttfamily]
public int sortaSum(int a, int b){
    if (a + b <= 10 && a + b >= 20)
        return 20;
    else return a + b;
}
\end{lstlisting}
\end{minipage}
&
\begin{minipage}[t]{.32\linewidth}
\vspace{5pt}
\textit{Level 0:} Incorrect\\
\textit{Level 1:} \\
The program fails in the following case. \\
Input: a = 5, b = 6\\
Expected Output: 20\\
Your Output: 11\\
\textit{Level 2:} The code fails because it wrongly handles the condition for the sum of a and b. The condition should be if the sum is greater than or equal to 10 and less than or equal to 19, then return 20.\\
\textit{Level 3:} The code fails because it mistakenly checks if the sum of a and b is less than or equal to 10 and greater than or equal to 20. This condition will never be true as a number cannot be both less than or equal to 10 and greater than or equal to 20 at the same time.\\
\textit{Level 4:} Change the if condition from \texttt{if (a + b $<=$ 10 \&\& a + b $>=$ 20)} to \texttt{if (a + b $>=$ 10 \&\& a + b $<$ 20)}.
\end{minipage}\\\\
\hline
\end{tabular}
\vspace{5pt}
\caption{Prompt for feedback-ladder generation in GPT-4, and it's corresponding response.}\label{tab:prompt-example}
\end{table*}

\begin{table*}[t]

\centering

\begin{tabular}{l|r|r|r|r|r|r|r|r|r|r|r}
\hline
\textbf{Annotator} &
  \multicolumn{1}{r|}{\textbf{A}} &
  \multicolumn{1}{r|}{\textbf{B}} &
  \multicolumn{1}{r|}{\textbf{C}} &
  \multicolumn{1}{r|}{\textbf{D}} &
  \multicolumn{1}{r|}{\textbf{E}} &
  \multicolumn{1}{r|}{\textbf{F}} &
  \multicolumn{1}{r|}{\textbf{G}} &
  \multicolumn{1}{r|}{\textbf{H}} &
  \multicolumn{1}{r|}{\textbf{I}} &
  \multicolumn{1}{r|}{\textbf{J}} &
  \multicolumn{1}{r}{\textbf{Avg}} \\ \hline
\textbf{A}   & 1.00 & 0.65 & 0.11  & 0.04  & 0.21  & 0.59  & 0.65 & 0.18  & 0.48  & 0.18 & 0.41          \\
\textbf{B}   & 0.65 & 1.00 & 0.20  & 0.13  & 0.07  & 0.53  & 0.48 & 0.14  & 0.40  & 0.27 & 0.39          \\
\textbf{C}   & 0.11 & 0.20 & 1.00  & -0.05 & -0.09 & 0.03  & 0.12 & 0.37  & 0.00  & 0.48 & 0.22          \\
\textbf{D}   & 0.04 & 0.13 & -0.05 & 1.00  & 0.11  & -0.10 & 0.14 & -0.06 & -0.06 & 0.09 & 0.12          \\
\textbf{E}   & 0.21 & 0.07 & -0.09 & 0.11  & 1.00  & 0.17  & 0.09 & 0.06  & 0.03  & 0.10 & 0.18          \\
\textbf{F}   & 0.59 & 0.53 & 0.03  & -0.10 & 0.17  & 1.00  & 0.39 & 0.23  & 0.43  & 0.19 & 0.35          \\
\textbf{G}   & 0.65 & 0.48 & 0.12  & 0.14  & 0.09  & 0.39  & 1.00 & 0.01  & 0.37  & 0.28 & 0.36          \\
\textbf{H}   & 0.18 & 0.14 & 0.37  & -0.06 & 0.06  & 0.23  & 0.01 & 1.00  & 0.14  & 0.25 & 0.23          \\
\textbf{I}   & 0.48 & 0.40 & 0.00  & -0.06 & 0.03  & 0.43  & 0.37 & 0.14  & 1.00  & 0.05 & 0.29          \\
\textbf{J}   & 0.18 & 0.27 & 0.48  & 0.09  & 0.10  & 0.19  & 0.28 & 0.25  & 0.05  & 1.00 & 0.29          \\ \hline
\textbf{Avg} & 0.41 & 0.39 & 0.22  & 0.12  & 0.18  & 0.35  & 0.36 & 0.23  & 0.29  & 0.29 & \textbf{0.28} \\ \hline
\end{tabular}%
\vspace{5pt}
\caption{Inter-rater agreement between annotators measured in Pearson correlation coefficient (PCC) values.} \label{tab:IRA}
\end{table*}

\section{Introduction}\label{sec1}

One of the primary ways humans learn is through feedback. In a class, where a teacher has to interact with numerous students, manually providing appropriate feedback for each student can be time-consuming. However, \cite{Johnston2015TheEO} shows the effect of immediate, just-in-time feedback on student learning, and \cite{Kochmar2020AutomatedPF} shows the effect of personalized feedback on student learning. The current practice of human feedback through teachers and teaching assistants can often be one-size-fits-all and not-on-time for students. Intelligent tutoring systems and online learning platforms have the potential to enable automatic feedback and scale up teacher effort, especially in the setting of learning programming. Due to the highly structured nature of programming languages, there exists a body of work on feedback generation for students in programming tasks, and have had some significant success.

Automated feedback methods in programming scenarios can be grouped by their forms: \emph{edit-based} feedback on code edits that can fix bugs in student code, and \emph{natural language}-based feedback that explains student mistakes, ask Socratic questions, or provide suggestions, in a conversational manner. On edit-based feedback, earlier works either focused mostly on pointing out syntax errors in student code or providing edit-based suggestions on how to fix student-written buggy code. \cite{paassen2017continuous} provides next-step hints in the edit-distance space and formulates a mathematical framework for edit-based hint generation policies. \cite{piech2015learning} introduces a neural network-based method to encode programs as a linear mapping from an embedded pre-condition space to an embedded post-condition space and proposes an algorithm for feedback at scale using these linear maps as features. On natural language-based feedback, recent advances in large language models (LLMs) enable us to wrap fluent utterances around any feedback in a conversational scenario, enabled by their generative capabilities and knowledge of code acquired via pre-training, which we detail below. 

In addition to forms, approaches for automated feedback generation can also be grouped by the type of student errors to address: \emph{syntax errors} and \emph{logical errors}. Syntax errors have been studied rigorously, especially since the rise of transformer-based generative models. \cite{yasunaga2020graph} uses a graph-based, self-supervised learning paradigm for program repair from diagnostic feedback (compiler messages). Break-It-Fix-It \cite{yasunaga2021break} uses a critic (e.g., compiler) and a fixer to learn realistic errors humans make in their code. These models can then be used to generate good code from bad ones using the fixer model and vice versa using the breaker model. PyFiXV \cite{phung2023syntax} uses a novel run-time validation mechanism to decide whether the generated feedback is suitable to share with the student. \cite{leinonen2023syntax} uses LLMs to enhance programming error messages with explanations of the errors and suggestions on how to fix them. All these methods though work well, can not address logical errors in programs. 

Automated feedback for logical errors in programming education has also been the focus of several recent works. \cite{kumar2024testcase} generates a set of test cases that can be used to validity of the student submission on the given problem. This is one of the most basic form of logical error feedback a system can generate to guide students' debugging their submission. Though it is effective for advanced students, beginner students may face significant challenge with this sort of feedback. \cite{kumar2024socratic} is also can be quite effective in helping advanced students since it generates a series of Socratic questions to guide students. These questions are  meant to spark self-realizations in the student which is highly effective in students' learning. However, this type of feedback is hard to follow for the beginner students. \cite{phung2023benchmark} systematically evaluates two LLMs, ChatGPT and GPT-4, and compares their performance to that of human tutors for a variety of scenarios in programming education such as program repair, hint generation, grading feedback, pair programming, contextualized explanation, and task synthesis. Their method produces a concise, single-sentence hint for the bug in the program, which is evaluated using human annotators. The experiment was only conducted for 5 simple programming problems and only 2 human annotators were involved. Similarly, \cite{phung2023hintgen} investigates the role of LLMs in providing human tutor-style programming hints to help students resolve errors in their buggy programs and develops a novel technique, GPT4Hints-GPT3.5Val. It uses a similar prompting strategy to the one in \cite{phung2023benchmark} to generate hints augmented with the failing test case and the fixed program. An extra validation step using GPT3.5 is used to validate the generated feedback. This approach requires preexisting test cases to find the failing test case for the current buggy program. \cite{pankiewicz2023feedback} conducts an experimental study where they used GPT3.5 to generate real-time feedback for the students and measure the effect of the feedback. All three of these methods lack personalization of feedback, only generate feedback of one type, and fall short in quality evaluation experiments for the generated feedback. 

In this paper, we explore the capability of GPT-4, a well-known LLM, in generating ``feedback-ladder'', a multilevel feedback cascade for a single student-written buggy problem. \cite{sheese2023studentqueries} shows that different students need different types of help when solving a problem. \cite{gao2022you} shows that at various stages of learning, students' help-seeking behavior changes. Inspired by these findings, we expect that generating feedback at \emph{multiple levels}, forming a ladder, will be helpful to students.  Feedback at a higher, coarse-grained level supports students conceptually and does not give away the solution, which is better for students with higher abilities and offers them more opportunities to think for themselves. Feedback at a lower, fine-grained level supports students more specifically and directly points out where their mistakes, which is better for students with lower abilities and prevents them from getting stuck. Therefore, teachers can use the feedback-ladder for personalized feedback by showing different levels to students with different needs, possibly by starting at a higher level and moving down if the student still struggles.
%
Our work is driven by a key research question: \textbf{Can we automatically generate relevant and effective feedback-ladders to address students' logical errors in introductory programming assignments?}


\subsubsection*{Contributions} In this paper, we detail a method for automatically generating feedback-ladders for logical errors in programming assignments using GPT-4. Our contributions are as follows:
\begin{itemize}
    \item We propose an LLM-based method that automatically generates feedback-ladders to address logical errors in programming assignments. 
    \item We conduct a user study that consists of annotators who are students, instructors, or researchers in programming education to evaluate the quality of the generated feedback-ladder.
\end{itemize}

We observe in our user study that each level has a similar effectiveness score across different questions, though higher levels have lower scores than the lower levels. We also observe that it is harder to generate effective feedback-ladder for higher-scoring submissions than lower-scoring submissions.

\section{Methodology}

\subsection{Feedback-ladder} \label{sec:feedaback-ladder}
We define feedback-ladder as a set of varying levels of feedback for a student-submitted, incorrect (including partially correct) program instance, $P_s$. Each level in the ladder corresponds to how much information is contained in the hint shown to the student. With increasing levels of feedback in the ladder, the learning gain of the student diminishes \cite{miwa2013stoic}. The feedback levels are defined as:\\
\textbf{Level 0 (Yes/No)}: The feedback only indicates whether $P_s$ is correct or not. If a student can utilize this level of feedback, then they are going to learn the most. \\
\textbf{Level 1 (Test Case)}: The feedback generates a test case $C_{fail}$, that when given to $P_s$ will give the wrong answer. It consists of the input for $C_{fail}$, the expected correct answer according to the problem, and what $P_s$ gives as the output. Utilizing this level teaches students debugging skills, which are hard to learn as beginners, and very conducive to their learning. \\
\textbf{Level 2 (Hint)}: The feedback generates a high-level description of the logical error in the code that is responsible for the failure. This level of feedback should focus on the conceptual error that a student might have and must avoid any suggestion regarding editing the code. Feedback on this level has the best balance for a student to correct the mistake, as well as achieve learning gains. \\
\textbf{Level 3 (Location)}: This level points out the lines at which the mistake occurred in the program. The feedback should refrain from explicitly mentioning the actual mistake or any edit suggestions. A student with sufficient programming knowledge should be able to correct the mistake with feedback on this level. \\
\textbf{Level 4 (Edit)}: At this level, the feedback generates edits that turn the student-submitted buggy program, $P_s$, into a correct program. The feedback should modify the existing structure of $P_s$ and produce something as close as possible while showing only the edits instead of the whole corrected program. The learning gain for this level is minimal, though it can be helpful for students who are absolute beginners. 

\subsection{Generating Feedback-Ladder}
We use the definitions of different levels in feedback-ladder in the prompts for GPT-4 to generate feedback in each level for a student-submitted buggy program. The prompt includes these definitions plus both the problem statement ($Q$) and the buggy program $P_s$ itself. We use the openAI API to prompt GPT-4. Table \ref{tab:prompt-example} shows an example prompt to generate the feedback-ladder. Table \ref{tab:prompt-example} also shows the generated feedback-ladder for the prompt by GPT-4. While using the openAI API calls the parameters given are: model: `gpt-4', temperature = 0, max\_tokens = 1024. We have used the ChatCompletion API to send our request and the response is used as the feedback.


\begin{table*}[h!t]
\centering
\begin{tabular}{c}
\hline
\vspace{5pt}
\begin{minipage}{.985\linewidth}
\vspace{5pt}
\textbf{Problem Statement:} We'll say that a value is ``everywhere'' in an array if for every pair of adjacent elements in the array, at least one of the pair is that value. Return true if the given value is everywhere in the array.
\end{minipage}\\
\end{tabular}
\begin{tabular}{c||c}
\hline
\textbf{Low-Scoring Submission} & \textbf{High-Scoring Submission}\\ 
\hline
\begin{minipage}[t]{0.29\linewidth}
\textbf{Program:}
\begin{lstlisting}[breaklines, linewidth=\textwidth, basicstyle=\ttfamily]
public boolean isEverywhere(int[] nums, int val){
    for(int i = 0; i <= nums.length; i++){
         if(nums[i] == val && nums[i+1] == val){
             return true;   
        }
    }
    return false;
}
\end{lstlisting}

\textbf{Generated Feedback-Ladder}

\emph{Level 0:} The code is incorrect.

\emph{Level 1:} Test case:

Input: nums = [1, 2, 1, 3], val = 1

Expected output: true

Your output: false

\emph{Level 2:} The code fails because it checks if the current and the next element are both equal to the given value. However, the problem statement asks to check if at least one of the pair is that value.

\emph{Level 3:} You should modify the condition inside the if statement in your for loop.

\emph{Level 4:} Change your if condition to: \texttt{if(nums[i] != val \&\& nums[i+1] != val)}. This will check if neither of the pair is equal to the given value. If it finds such a pair, it will return false. If it doesn't find any such pair, it will return true after the loop.
\end{minipage}     
&  
\begin{minipage}[t]{0.67\linewidth}
\textbf{Program:}
\begin{lstlisting}[breaklines, linewidth=\textwidth, basicstyle=\ttfamily]
public boolean isEverywhere(int[] nums, int val){

    int a = 0;
    for (int i = 0; i < nums.length; i++){
        if (nums[i] == val || nums[i + 1] == val){
            a = a + a;
        }
        else{
            a = a + 1;
        }
    }
    return a == 0;
}
\end{lstlisting}
\textbf{Generated Feedback-Ladder}

\emph{Level 0:} The code is incorrect.

\emph{Level 1:}
Test case:

Input: nums = [1, 2, 1, 3], val = 1

Expected output: true

Your output: false

\emph{Level 2:} The code fails because it is not correctly checking if the value is everywhere in the array. The condition in the if statement is not correct and the increment of 'a' in the if and else blocks is also not correct.

\emph{Level 3:} You should modify the if condition in the for loop and the increment of 'a' in the if and else blocks.

\emph{Level 4:} Change the if condition to check if the current element or the next element is equal to the value. If it is, continue to the next iteration. If it is not, return false. At the end of the loop, return true. Here is how you can modify the code:
\begin{lstlisting}[breaklines, linewidth=\textwidth, basicstyle=\ttfamily]
for (int i = 0; i < nums.length - 1; i++){
    if (nums[i] == val || nums[i + 1] == val){
        continue;
    }
    else{
        return false;
    }
}
return true;
\end{lstlisting}
\end{minipage}\\
\hline
\end{tabular}
\vspace{5pt}
\caption{An example of generated feedback-ladders used in our experiment.}\label{tab:feedback-example}
\end{table*}
\section{Experiments}
\subsection{Dataset}
We use the dataset from the 2nd CSEDM Data Challenge, hereafter referred to as the CSEDM dataset \cite{csedm2019}. It is a college-level, publicly-available dataset with students' actual code submissions. It contains 246 college students' 46,825 full submissions on each of the 50 programming questions over an entire semester with rich textual information on problem statements and student code submissions and other relevant metadata such as the programming concepts involved in each question and all error messages returned by the compiler. From the full dataset, we take only program submissions that do not have syntax errors, since our work focuses on generating feedback for programs that have no syntax errors but contain logical errors in solving the given problem. Therefore, we only curate such submissions from the dataset with less-than-perfect scores. To keep the scale of the user study manageable, we handpicked 5 problems. These 5 problems require students to use arrays, strings, loops, and conditions in their programs. Therefore, these problems test a comprehensive set of programming knowledge for the students. These problems are also complex enough to have room for nuanced mistakes from the students. For each of these five problems, we select three different student-submitted programs (submissions), one each from the following three categories:\\
\textbf{Low-scoring submissions:} Submissions scoring less than 20\% in the test cases. These submissions have numerous errors and require significant edits to fix. \\
\textbf{Mid-scoring submissions:} Submissions scoring between 40\% and 60\% in the test cases. These submissions have fewer mistakes than the low-scoring submissions.\\
\textbf{High-scoring submissions:} Submissions scoring more than 80\% in the test cases. These are almost correct and often require a few edits to get the full score. However, finding the errors in these submissions is difficult since they often miss only a few corner test cases.  

\subsection{Experiment Design}
We conduct a user study to evaluate the quality of the generated feedback-ladder. The annotators for the user study are recruited from several universities, with different knowledge levels of programming: students, CS instructors, and AI researchers. There are a total of 10 annotators. The user study consists of several different phases, detailed below.\\\\
\textbf{Eligibility determination phase:} The purpose is to determine that the annotators have sufficient programming skills required by the study. First, the annotators are shown a simple program in Java. Second, they are given a set of inputs for the given program. Third, they are asked to determine the output of the program on the given set of inputs. Failing this disqualifies them from participating in the study. \\
\textbf{Calibration phase:} The purpose is to calibrate the evaluation objective among annotators to align with our expectations. First, the annotators are shown examples of different feedback-ladder. Each example contains a problem statement, a submitted program, and a generated feedback-ladder. Second, they are asked various questions about the shown examples. The questions are related to identifying the proper level of a feedback text, evaluating qualitative measures, etc. If the annotator got any of the answers wrong, we show which questions are wrong but do not reveal the answer. They try again and choose different answers for the same questions. This phase ends when the annotators answer all the questions correctly. \\
\textbf{Evaluation phase:} The purpose is to evaluate the quality of the generated feedback-ladder for 15 total programs, i.e., 5 chosen problems with 3 different submissions each. First, the annotator is shown a problem description. Then they are shown a low-scoring student-submitted program and the generated feedback-ladder for it.
Second, they are asked to rate two different metrics of each of the feedback on a 5-point Likert scale, which we detail below. This repeats two more times for the mid and high-scoring submissions. The whole thing repeats for four more problems.\\
\textbf{Evaluation Metrics:} We ask study annotators to rate the feedback of each level on the feedback-ladder on \textit{Relevance} and \textit{Effectiveness}. The \textit{relevance} of feedback for a particular level is based on how well the feedback matches its level definition. 5 corresponds to a perfect match and 1 corresponds to no match. The \textit{effectiveness} of a generated feedback for a particular level is how effectively it can help a student, according to the annotator's judgment. 5 corresponds to highly effective feedback that should help the student fix their bug and 1 corresponds to feedback that does not help the student or may confuse them. 

\section{Results and Discussion}

\subsection{Inter-Rater Agreement} Table \ref{tab:IRA} shows the inter-rater agreement between pairs of annotators in our study, measured using Pearson correlation coefficient (PCC) \cite{cohen2009pearson}, with average PCC values is reported at the end of the row and column for the corresponding annotators. We see that the overall average of all the pairwise PCC values is 0.28, which indicates moderate agreement across annotators.



\begin{figure}[h]
    \centering
    \begin{subfigure}{\linewidth}
        \centering
        \includegraphics[scale=.33]{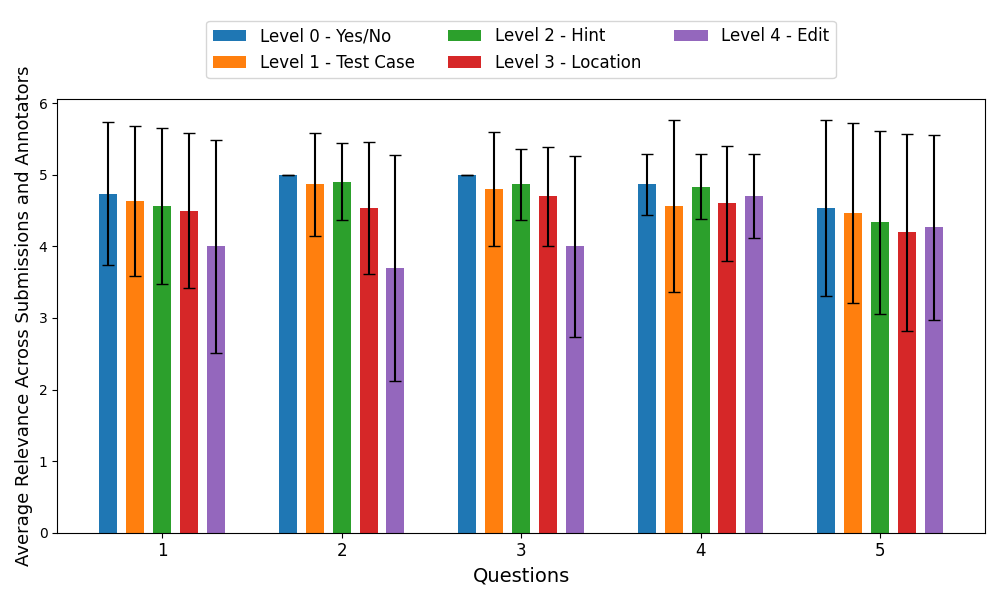}
        \caption{Relevance}
    \end{subfigure}
    \begin{subfigure}{\linewidth}
        \centering
        \includegraphics[scale=.33]{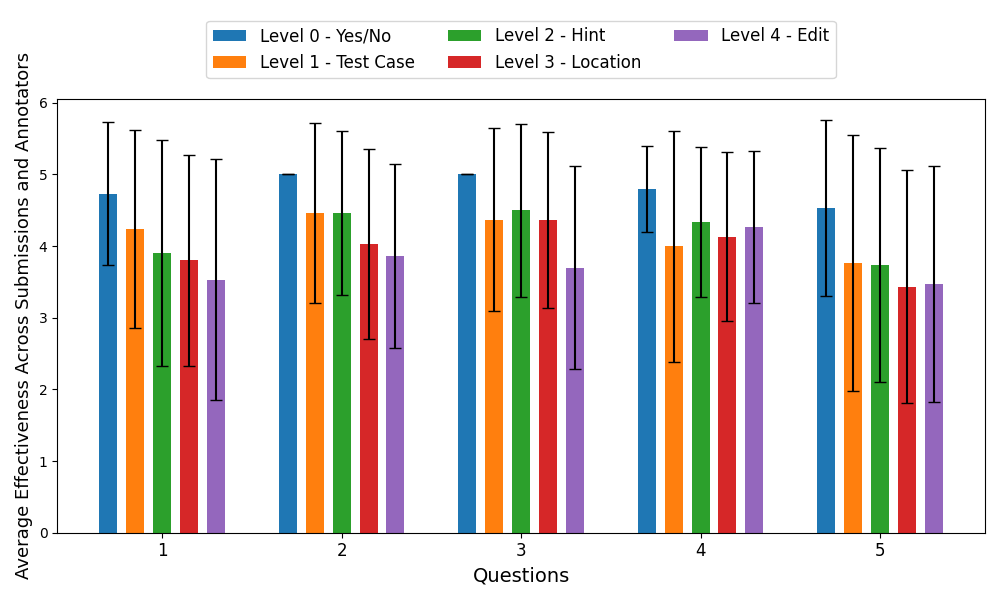}
        \caption{Effectiveness}
    \end{subfigure}
    \caption{Relevance and Effectiveness score for each question, reported for each level - averaged over all three submissions and all annotators. Error margin is standard deviation.}
    \label{fig:questionwisestats}
\end{figure}

    

\begin{figure}[h]
    \centering
    \begin{subfigure}{\linewidth}
    \centering
      \includegraphics[scale=.33]{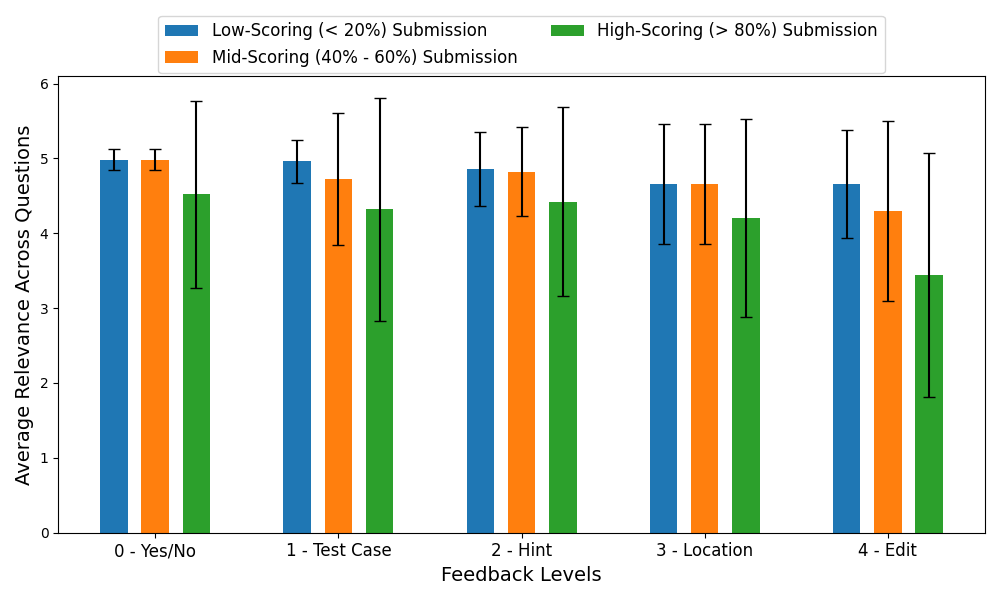}
      \caption{Relevance}
    \end{subfigure}
    \begin{subfigure}{\linewidth}
        \centering
        \includegraphics[scale=.33]{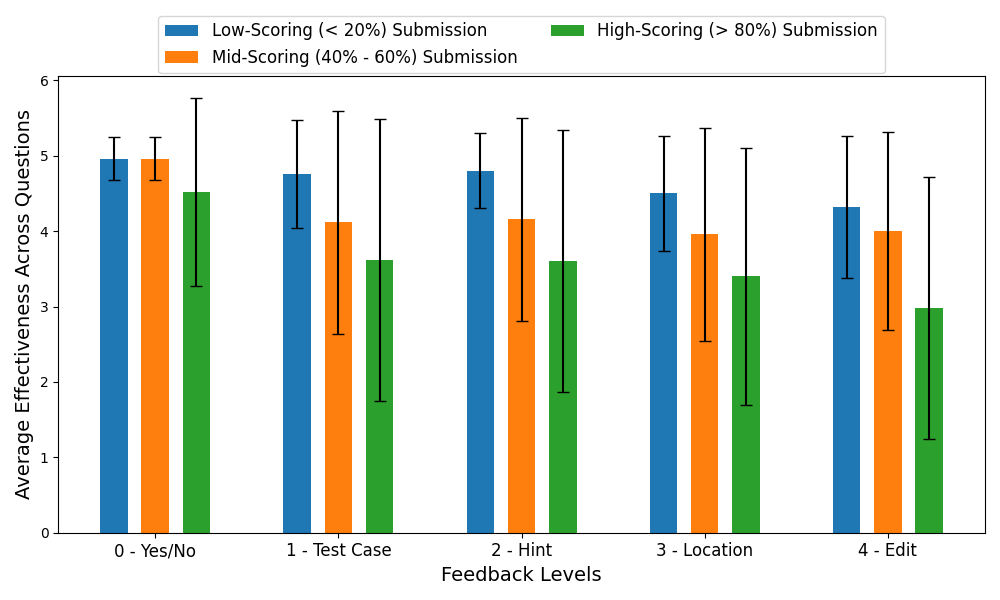}  
        \caption{Effectiveness}
    \end{subfigure}
    
    \caption{Relevance and Effectiveness score for each level, reported for all three submissions - averaged over all problems and annotators. Error margin is standard deviation.}
    \label{fig:levelwisestats}
\end{figure}

\subsection{Analysis of Generated Feedback-Ladder} Table \ref{tab:feedback-example} shows an example of generated feedback-ladders we used in the experiment. In the low-scoring submission, the mistake occurs in the condition of the \texttt{if} statement. The student checks whether the current element and the next element are both equal to \texttt{val}, despite the problem requiring them to check if one of them is equal to \texttt{val}. In Table \ref{tab:feedback-example}, we see that levels 0 to 3 address this error effectively according to each level's definition. In the case of level 4, the method suggests edits that correctly solve the mistake but require more edits than necessary, since changing the \texttt{\&\&} to \texttt{||} in the condition is sufficient to correct the mistake. The feedback-ladder does not address the potential ``Array Index Out of Bound'' mistake in the code either; however, it is desirable since the feedback-ladder is supposed to focus on a single mistake at a time. 

In the high-scoring submission, the mistake in the program is the potential ``Array Index Out of Bound'' problem in the \texttt{if} condition. All the feedback messages in levels 1 to 4 fail to address this mistake. In addition, level 2 and 3 feedback point out non-existing, hallucinated mistakes in the student code. The level 4 feedback corrects the program but changes the main structure of the student-written program, which is absolutely undesirable since the resulting feedback is hardly relatable to the student. Simply changing the for loop condition to \texttt{i < nums.length - 1} is sufficient to fix the program. This example encapsulate some trends we see in the study results that we further detail below. 

\subsection{Trends Found in the User Study}
Figure \ref{fig:questionwisestats} shows the trend for how annotators' relevance and effectiveness ratings change across questions. There are five sets of bars, each for one question in our study. Each set contains the average score for each of the levels represented using a single bar, together with standard deviations. The scores are averaged over all scores from all the annotators and all three submissions for each question. Figure \ref{fig:levelwisestats} shows the trend on the same evaluation metrics across different feedback levels. We can make two key observations from these results:\\
\textbf{Quality of feedback for each level is consistent across questions.} We see that the quality of feedback for lower levels is better than for higher levels in 4 out of 5 questions. In terms of relevance, we see that our method can generate highly relevant lower-level feedback. However, for higher-level feedback, the relevance diminishes. Upon manually inspection, we observe that as the level increases, the feedback diverges from the level definition more and more. For example, level 4 feedback often has the whole program listed or contains a brand new program in the generated feedback-ladder, which violates the definition of level 4 feedback. On the other hand, the generated feedback is almost perfectly relevant at level 1 and level 2 since these levels require highly structured feedback. Overall, the method struggles to maintain relevance for levels 2, 3, and 4 a little bit more than levels 0 or 1. In terms of effectiveness, we see a similar trend as the one for relevance, except that the effectiveness at higher levels is much lower. In the annotators' opinion, listing whole programs or replacing them with a new program has a more negative impact on effectiveness compared to relevance. An exception, though, can be found between levels 1 and 2, where even though level 1 has low effectiveness, level 2 has higher effectiveness for questions 2, 3, and 4. This means that LLM-generated hints have higher quality than generated test cases. \\\\
\textbf{Quality of feedback is higher for low-scoring submissions than high-scoring submissions.} Among submissions with different scores (low-scoring: $<20\%$, mid-scoring: $40\%-60\%$, high-scoring: $>80\%$), we observe a clear trend that GPT-4 struggles to generate good feedback for higher-scoring submissions. This result is likely because finding a mistake in low-scoring submissions is much easier than doing so for higher-scoring submissions; high-scoring submissions are almost correct and often contain tiny mistakes that fail one or two corner test cases. Even human experts struggle to find these mistakes. We also find evidence of these situations upon manual inspection of the annotator ratings: The high-scoring submissions for Q1 and Q5 were so close to being correct that two of our annotators mistakenly believed they were correct. Therefore, they rated the generated feedback-ladder as irrelevant and ineffective at all levels. 

\section{Conclusion and Future Works}
In this paper, we explored large language models, specifically GPT-4's capability to generate feedback-ladders for introductory programming assignments. An experiment using a user study suggests that despite showing some promise, this approach has some limitations in generating highly effective feedback, especially at higher levels when suggesting providing more specific feedback.

The current method of feedback-ladder generation can be a useful tool for teachers. Based on the different knowledge levels of different students, teachers can choose the appropriate feedback level to intervene. Though the feedback for high-scoring submissions may not be highly effective, the method may still be able to reduce the workload of teachers since it can effectively address mistakes in low-scoring submissions. Even in cases where the feedback is moderately effective, giving teachers a starting point can potentially reduce the time it takes for them to write feedback, especially compared to writing it from scratch. 

For future work, we can think about multiple options. First, we can run a large-scale, real-classroom study with real-time interventions. In this way, we can measure the effectiveness of the generated feedback-ladder using actual learning gains of the students using the feedback. Second, we can train models that can personalize the feedback by choosing which level of feedback is appropriate for which student. Lastly, we can train a dedicated LLM for the task of feedback-ladder generation using expert feedback as using GPT-4 can be costly both in terms of time and money.
\bibliographystyle{splncs04}
\bibliography{sn-bibliography}


\end{document}